# E-Commerce Inpainting with Mask Guidance in Controlnet for Reducing Overcompletion

Guandong Li[1*]

1.*Suning, Xuanwu, Nanjing, 210042, Jiangsu, China.

*Corresponding author(s). E-mail(s): leeguandon@gmail.com

*Abstract*: E-commerce image generation has always been one of the core demands in the e-commerce field. The goal is to restore the missing background that matches the main product given. In the post-AIGC era, diffusion models are primarily used to generate product images, achieving impressive results. This paper systematically analyzes and addresses a core pain point in diffusion model generation: overcompletion, which refers to the difficulty in maintaining product features. We propose two solutions: 1. Using an instance mask fine-tuned inpainting model to mitigate this phenomenon; 2. Adopting a train-free mask guidance approach, which incorporates refined product masks as constraints when combining ControlNet and UNet to generate the main product, thereby avoiding overcompletion of the product. Our method has achieved promising results in practical applications and we hope it can serve as an inspiring technical report in this field.

1. Introduction

The generation of product images based on diffusion models has already produced significant benefits in the e-commerce field. In the pre-AIGC era, our work primarily focused on template analysis[1], generation[2], synthesis, and deployment[3]. However, traditional creative production platforms still lack the capability to directly generate reasonable images with light and shadow effects based on product images. This task often requires the use of templates. Although the generated images possess layer attribute information, they are essentially still programmatic creativity based on search and replace, which can be used in scenarios with strict image format requirements, such as during major promotions or for main images, but lack appeal in marketing contexts. With the emergence of AIGC, particularly stable diffusion[4], the powerful generative capabilities of text-to-image diffusion models can typically generate logically coherent images with appropriate lighting based directly on product images[5]. This presents a vast range of applications and potential in marketing scenarios for product images.

The synthesis of e-commerce images requires a complete workflow, typically including: 1. Acquiring the necessary product image; 2. Writing prompts and generating content; 3. Using a pipeline of inpainting, LoRA[6], and ControlNet[7] for image generation; 4. Upscaling to enhance image resolution. In the first step, users often upload various images that include the products to be generated, rather than images that have been cut out. Therefore, we need to provide interactive tools like hqsam[8] for image segmentation, and in some categories, direct cutouts may be necessary. In the second step, it is essential to describe the input product. We typically use BLIP[9] to generate a few brief descriptive sentences and utilize CLIP to extract image features that match the text features from a predefined vocabulary. The extracted tags are used to enrich the product image prompts. The third step is the most critical, as the core of product image generation lies in inpainting. Inpainting controls the main product image, allowing for the generation of scenes that match the product image. Combined with the underlying e-commerce text-to-image model and corresponding scene LoRA, specific scene images can be produced. ControlNet serves to better constrain the product image. This paper will focus on analyzing inpainting and propose the use of mask guidance technology in ControlNet to address the challenges in product image generation, particularly the issue of overcompletion. The fourth step involves expanding the image as needed. Product image generation can be roughly divided into these four steps.

The e-commerce image generation process described above faces several challenges. For instance, the synthesized images may not match the background well, resulting in a stiff appearance. After synthesis, the product images may exhibit glowing or blurring at the edges during inpainting, and the background may be excessively blurred. Among these issues, the most difficult to handle is overcompletion, which can lead to difficulties in maintaining the features of the main product and may result in strange branches or shapes emerging from those features. In the workflow, optimizing the image segmentation is necessary to minimize irregular or jagged textures during inpainting, although this is not detailed in this paper. The core focus of this paper is on the inpainting module and the train-free mask guidance module in ControlNet[10]. We evaluate several existing inpainting methods and identify that the root cause of overcompletion lies in the use of a large amount of random mask data during training. Therefore, we propose a two-stage fine-tuning scheme based on instance masks. In this approach, we do not use the mask module to guide inpainting, thereby reducing its pressure. Instead, we apply the train-free mask guidance module in ControlNet, where the pre-obtained product image masks are refined and fused with the output from the UNet layer in ControlNet. This dual approach aims to reduce the issue of overcompletion effectively.

Our contributions include: 1. Proposing a complete workflow for e-commerce image generation; 2. Discussing methods for e-commerce image inpainting; 3. Addressing the issue of overcompletion by optimizing inpainting methods and introducing train-free ControlNet mask guidance.

2. Related works

2.1 Inpainting method

Inpainting is the core of the entire e-commerce image redrawing process. Optimizing the inpainting solution can resolve the majority of issues related to poor image generation.

Standard inpainting: blend latent diffusion[11] can achieve inpainting effects based on the text-to-image (T2I) model, combined with image-to-image (img2img) techniques, allowing the model to generate only in specified areas while referencing other regions during the generation process. The denoise parameter indicates the intensity of reconstruction, and the generation scheme is based on blend mask fusion, adjusted through sampling strategies. However, the limited perception of the mask boundary regions can lead to incoherent repair results.

Soft inpainting[12] is a method used in web UI that achieves a smooth transition between the masked and non-masked areas by utilizing a soft (grayscale) mask. This differs from standard repairs that use black-and-white masks. In the transition region where the mask is between black and white, the latent images of the inpainted content and the original content blend during sampling to achieve a seamless transition.

Inpainting model: The inpainting model cannot reuse the T2I model, as the in_channels of the UNet changes from 4 to 9 (4+4+1). The first 4 channels represent the sampled noise, the second 4 channels represent the masked image, and the last 1 channel is the binary mask. This is a model that requires training. Aside from the change in channel numbers, the subsequent sampling process remains the same as usual. After sampling, the VAE decodes to obtain the image. Currently, in addition to the inpainting models provided by the official SD 1.5 and SDXL[13], there are several effective inpainting models available in the community, many of which also utilize model fusion techniques to generate results.

ControlNet inpainting: ControlNet provides an inpainting mode, which allows the input of masked areas to control image generation. It can be used in combination with the T2I model, and in the ControlNet branch, only the masked image needs to be input.

Fooocus inpainting[14]: It can convert any T2I model into inpainting mode. It includes a fooocus_inpaint_head, which compresses the 9 channels into a smaller convolutional network with 4 channels. The standard UNet has 4 inputs, while the inpainting model has 9 channels. It also includes an inpaint model, where the head is responsible for channel compression, and the model enhances the redrawing capability of any model. Since the author has not disclosed the method, it is speculated that the training approach may involve adding a similarly initialized UNet for training, freezing the original UNet, and allowing only the new UNet to train, maximizing the learning of redrawing capability, and then merging the weights during forward inference.

BrushNet[15]: BrushNet is a specialized inpainting model that inserts masked image features into a pre-trained diffusion network through an additional branch, separating the extraction of masked image features from the generation process. When processing masked image features, BrushNet uses the weights of the pre-trained diffusion model and does not input text embeddings, ensuring that the additional branch only considers pure image information. The features from BrushNet are inserted layer by layer into the frozen diffusion model, achieving dense hierarchical control over each pixel, using zero convolution to connect the locked model and the trainable BrushNet. BrushNet is the chosen inpainting architecture for this paper.

2.2 Controlnet mask guidance

Our idea in ControlNet mask guidance comes from IP-Adapter masking[16]. IP-Adapter masking can control different reference areas in the generated image through a mask. In ControlNet, we multiply the feature maps produced by the encoder UNet of ControlNet by an additional mask, hoping to fix the generation of the masked areas.

3. Method

In this section, we will detail the proposed Ecommerce Inpainting framework and how to add train-free mask guidance in ControlNet, as shown in Figure 1. Typically, generating e-commerce images only requires selecting an appropriate inpainting method. However, in practice, relying solely on inpainting has many shortcomings, such as poor control over lighting and shadows, and overly rigid blending. Therefore, our approach includes both inpainting and ControlNet. When considering the use of mask guidance to control the generated product image's main subject, we aim to fully leverage the generative capabilities of the inpainting model while minimizing the impact of overly strong mask constraints. For instance, the soft inpainting method may produce smoother transitions at the boundaries between masked and non-masked areas, possibly due to the crude binary constraints of the mask. Thus, we do not consider introducing mask guidance during the inpainting process; instead, we implement it in ControlNet, where applying mask constraints to the feature maps of the UNet encoder is more reasonable.

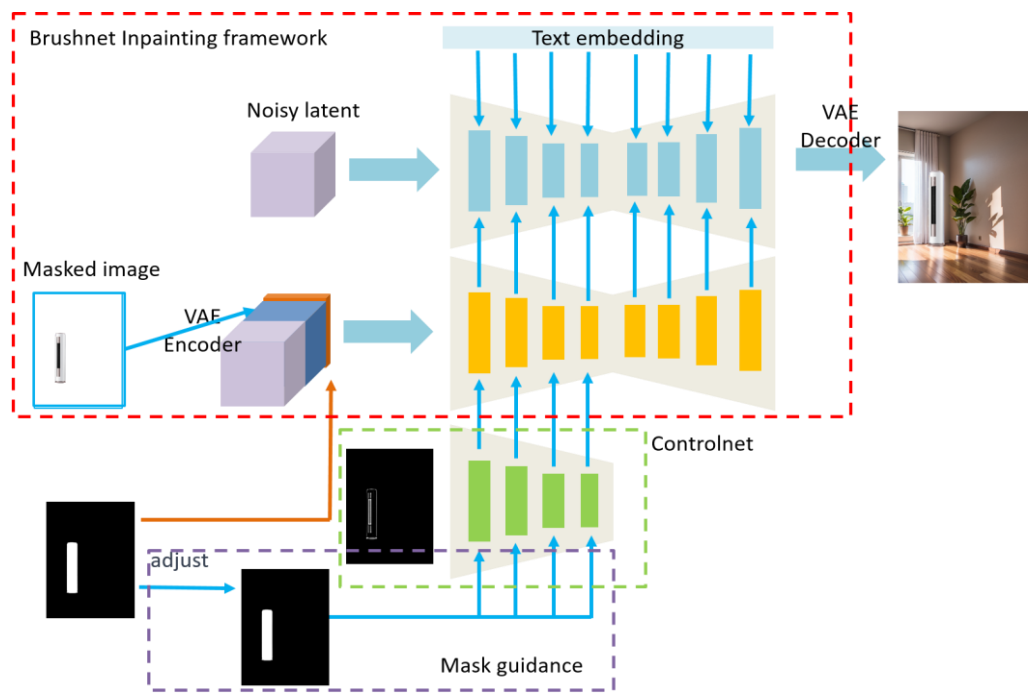

Figure 1: E-commerce image generation framework guided by mask guidance in ControlNet.

3.1 Ecomerce inpainting

3.1.1 How to choose an inpainting architecture?

Faced with numerous inpainting methods, how should we choose our foundational inpainting framework and how can we perform secondary optimization based on that framework? We hope that the inpainting framework can couple well with text-to-image (T2I) methods, allowing for the use of more personalized style images. From this perspective, the standard/soft inpainting, Fooocus inpainting, ControlNet inpainting, and BrushNet we mentioned all meet the requirements, indicating that this is also a trend in the current development of inpainting. However, the 9-channel inpainting model requires additional training for each specific scenario. Standard/soft inpainting is not considered, as it performs poorly in e-commerce image scenarios. Fooocus inpainting shows relatively average performance in e-commerce contexts, and since the authors have not publicly disclosed its specific implementation logic, it is not user-friendly for subsequent updates based on vertical data. ControlNet inpainting and BrushNet essentially do not differ much; both have an additional separate branch for extracting mask features. However, compared to BrushNet, ControlNet inpainting's ability to extract information and integrate into UNet is relatively insufficient, and inpainting requires pixel-level constraints. Therefore, we ultimately chose BrushNet as our foundational architecture. Although the coupled inpainting module is our final choice, there are also some good 9-channel inpainting models in the community, such as the Realistic Vision inpainting model, which performs very well. However, its inpainting module weights may also be fused, so experienced testing of its effectiveness is necessary. Additionally, BrushNet has its drawbacks; the original BrushNet was trained on the Realistic Vision base model. In fact, when we change the base model, its performance may degrade to some extent, and its generalization ability may also be somewhat limited. Therefore, in practical scenarios, a balance needs to be struck between compatibility and effectiveness. We also want to emphasize that the base

T2I model is very important; if specific scene inpainting is desired, matching its LoRA is equally important.

Our inpainting architecture adopts BrushNet, which has a complete additional branch aligned with UNet. The input to the additional branch includes the noise latent vector, the latent vector of the mask image, and the downsampled mask. The inputs are concatenated into the model, where the noise latent vector provides information during the generation process, helping BrushNet enhance the semantic consistency of the mask image features. The latent vector of the mask image is extracted from the mask image using a VAE, aligning with the data distribution of the pre-trained UNet. The mask is downsampled using cubic interpolation to ensure consistent dimensions. When processing the mask image features, BrushNet uses the weights of a pre-trained diffusion model, and the additional branch does not embed a text branch, ensuring that the additional branch only considers pure image information. The features from BrushNet are inserted layer by layer into the frozen diffusion model, achieving dense hierarchical control over each pixel, using zero convolution to connect the locked model and the trainable BrushNet.

3.1.2 How to optimize excessive completion at the inpainting level?

One important reason for excessive completion in product image generation is that most general inpainting modules construct masks using random masks, which are then used to segment the foreground and background for training. Objects in the image can easily be truncated by these masks, leading the trained model to tend to complete the appearance and shape of the objects based on associations. To address this issue, we fine-tuned BrushNet using instance masks, which are primarily sourced from e-commerce image scenarios and some salient object detection datasets.

3.2 Controlnet mask guidance

Training instance masks in the inpainting pipeline can alleviate the phenomenon of excessive completion. In addition, we also want to explore whether there are train-free methods to optimize this. Since inpainting is the core link of generation, we do not want to impose restrictions on the generative capabilities of inpainting. Therefore, we added mask guidance in the ControlNet pipeline to constrain the generated product image subjects. In the IP-Adapter masking pipeline, we can control the generated image by specifying regions. Binary masks specify which portion of the output image should be assigned to an IP-Adapter, and we can introduce binary masks as attention masks on the cross-attention side, similar to region prompts. Inspired by this idea in ControlNet, we can input a mask as a constraint, multiplying a mask at each layer of the encoder UNet branch in ControlNet, and performing element-wise multiplication with the output of each layer in ControlNet. The resulting output is then fed into the decoder of UNet, effectively isolating the influence of the background. Additionally, we perform some extra preprocessing on the masks: we first apply a closing operation to fill small holes, then an opening operation to remove small noise, and finally a dilation operation to enhance the edges. This effectively refines the edges of the mask while preserving its overall shape. Our mask guidance operation is applied to 13 layers of the encoder in ControlNet's UNet, with sizes of 64x64, 32x32, 16x16, and 8x8 across the four levels. Our ControlNet mask guidance can be applied in modes such as Canny, OpenPose, and LineArt. In ControlNet, these models have pure black as empty space, meaning no control will be applied to pure black regions. However, for other models (tile, depth), there is no way to specify the control region, as pure black regions have their own semantics for these models. Of course, ControlNet mask guidance is not limited to the BrushNet framework mentioned in this article;

in fact, it can be integrated with any inpainting framework and achieve good results.

4. Results

As shown in Figure 2, our method demonstrates a good solution to the excessive completion problem, as well as the visual effects of the generated e-commerce images.

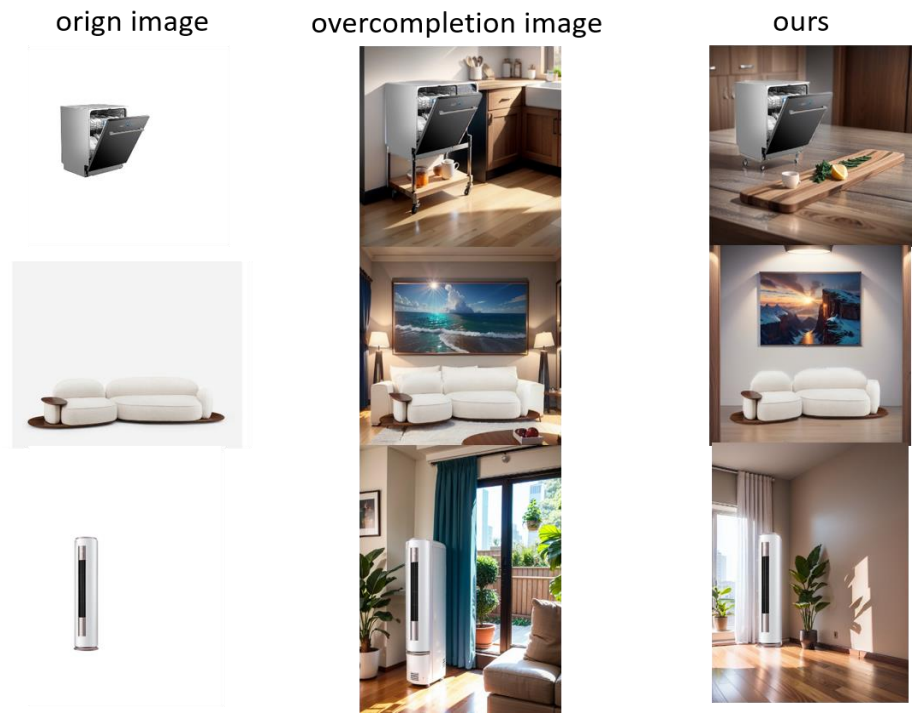

Figure 2: Model results

5. conclusion

This paper discusses the causes of excessive completion in e-commerce product image generation and presents a series of solutions to address this issue. We chose a BrushNet-based architecture for the inpainting method in e-commerce scenarios, fine-tuned with instance masks, and introduced a train-free approach based on ControlNet's mask guidance. By incorporating product subject mask constraints during the integration of ControlNet and UNet, we effectively generated product subjects in the UNet decoder. Mask guidance can be combined with any inpainting framework. These combinations successfully resolve the excessive completion problem. Furthermore, this paper introduces a comprehensive solution for e-commerce image generation, hoping that it can serve as an inspiring technical report in the field of e-commerce image generation.